\documentclass[12pt]{article}

\usepackage{graphicx}
\usepackage[margin=1in]{geometry}
\usepackage{amsmath}
\usepackage{amssymb}
\usepackage{subfigure}

\begin{document}
\title{Modeling the power consumption of a Wifibot and
studying the role of communication cost in operation time}
\author{Dimitrios Zorbas and Tahiry Razafindralambo\\
\small \texttt{jim@students.cs.unipi.gr, Tahiry.Razafindralambo@inria.fr}\\
Inria Lille -- Nord Europe, France}
\date{}
\maketitle

\begin{abstract}
Mobile robots are becoming part of our every day living at home, work or
entertainment. Due to their limited power capabilities, the development
of new energy consumption models can lead to energy conservation and energy
efficient designs. In this paper, we carry out a number of experiments and
we focus on the motors power consumption of a specific robot called Wifibot.
Based on the experimentation results, we build models for different speed
and acceleration levels. We compare the motors power consumption to other
robot running modes. We, also, create a simple robot network scenario and we
investigate whether forwarding data through a closer node could lead to
longer operation times. We assess the effect energy capacity, traveling
distance and data rate on the operation time.
\end{abstract}


\section{Introduction}
\label{intro}

Nowadays, mobile robots are used in a variety of applications beginning from the most common
ones like vacuum-cleaners, pool-cleaners and drones and extending to more complex like
secure and rescue, and planet exploration \cite{siegwart2011introduction}.
The environmental monitoring is another application, where robots can be used to detect gas \cite{7005140},
air pollution \cite{ferri2010dustcart}, and mines \cite{999212}. The rescue mobile robot Quince
has been developed to resolve technical issues in case of a nuclear plant accident \cite{6106794}.
NASA's Mars Exploration Rovers is another robot which was used to discover Mars' surface and water
activity. In all these applications, the robots are powered by batteries and energy
conservation is critical for the longevity of the project.

In order to design power-efficient methods for mobile robots, it is important
to have a practical knowledge of the energy efficiency of the devices used. This paper provides
some models of the global power consumption based on the utilization of major robot power
consumers, such as the motors, the network adapter, the embedded controllers etc..
We carry out a set of experiments using
a mobile robotic platform called Wifibot \cite{wifibots}. It is a modern commercial mobile
robot which has been used in many research activities \cite{milan,katsikiotis,tahiry,potpet,quilez,tonneau}.
We believe that the results would be helpful for people who work with other types of robots, as
well as for researchers who work on the design of power efficient paths and scheduling techniques.

Our study has three major contributions.

We model the power consumption of several motors running modes using real experiments and we examine the role
of acceleration in power cost. As far as we know, this is the first work which models
the extra cost of acceleration in total power expense. The results show that this
cost can grow the total power consumption up to 66\% for the specific robot.

Second, we measure the power consumption utilizing other power consumers like the embedded system,
the processor, the wireless network interface, and the camera. The results show that only the
25\% of the total power is used for the motion, while the half is spent by the
embedded system. 

Third, we examine whether moving a robot to a new location and lowering the communication cost could
conserve energy and in long-term could lead to longer operation times. We assume a simple scenario
where a robot is used as relay which 
comes in between a robot with monitoring capabilities and a base station. The two robots adapt
their communication ranges to shorter radius and, thus, consume less energy. The 
results show minor improvements compared to the case where a direct long communication range is used.
Since these results are based on the specific robot, we present a number of generic conditions under which
the reduced communication cost could result in longer operation time.

\section{Related work}

Energy efficiency is an important issue in mobile robotics and providing accurate power consumption
models helps to the design of better algorithms to reduce the overall consumption. In this section, we
summarize the most recent works related to energy efficiency in mobile robots.

A significant amount of works has been done in the area of energy efficient trajectories design.
However, in these works, the motion consumption is considered as the unique power consumer in the
design of the algorithms. In this paper, we show that the embedded system, as well as the processing
and the communication consume significant amounts of energy. The authors in \cite{1302401}
focus on finding energy-efficient motion plans by determining velocities for predefined routes.
In \cite{1391019}, optimal paths are computed with respect to the energy consumption of the
robot. An efficient approximation algorithm that computes a path whose
cost is within a user-defined relative error ratio is presented. Similarly to this work, 
energy-efficient motion planning problems are considered in \cite{6027010,6265887} and \cite{6459035}.
In these works, A* algorithm is used to construct optimized trajectory plans which are computed
by dividing the path in segments; efficient angles and velocities are calculated for each segment.
Regarding the same works, theoretical and experimental power consumption values are taken into account
in \cite{6027010} and \cite{6459035,6265887} respectively. However, only motion power is taken into
account. Finally, an improved A* algorithm is presented in \cite{6896396} which significantly reduces
the energy expense in the authors examined scenario compared to the classic A* approach. 

Energy savings can be also achieved by properly scheduling the robot operations. For example,
in \cite{4107658}, an optimization method which controls both processor's frequency and motor's
speed is introduced to avoid collisions. In other works, the energy consumption for each operation
is modeled and parameterized as function of the operation execution time, and an energy-optimal
schedule is derived by solving a mixed-integer nonlinear programming problem \cite{5584686}.
Since robots consist of several embedded systems that we also meet in
a common computer, dynamically reconfiguring these systems to provide the required services
with the minimum number of active devices, could lead to energy conservation \cite{1507454}.
These dynamic power management techniques are summarized in \cite{845896,1213004} and they are
out of the scope of this paper.

Morales et al. \cite{Morales2009TRO} model the power consumption of a skid-steer tracked mobile
robot. The authors focus on modeling the power losses of dynamic friction since in this kind of
robots the contact area with the ground is high. 
Energy scavenging technologies for mobile robots are discussed in \cite{Vaussard1}. The authors 
assess the feasibility of using these technologies by conducting experiments with a set of robotic
vacuum cleaners. 
Vaussard et al. \cite{Vaussard2} study new technologies that could lead to energy reduction of
domestic mobile robots. They validate their study by analyzing the power consumption of seven
vacuum-cleaning robots.
In \cite{1507454} The authors present a case study of a
mobile robot energy consumption, named Pioneer 3DX, and they introduce some energy conservation
techniques. The consumption of other robot components is taken into account
in the experimental measurements. However, motor acceleration is not considered. As it is claimed
in this paper as well as in other works, the communication cost can be saved by placing intermediate
robots (nodes) between the robot which transfers data and a base station. It is assumed that
transmitting a high volume of data could save the motion cost to move closer to the transfer robot.
However, this assumption is not verified or quantified.

\section{Mobile robot architecture -- Wifibots}

A common mobile robot consists of some essential components, like the microcontroller, the motors,
the embedded computer, one or more sensors, and the batteries. In some robots the embedded computer
and the microcontroller refer to the same device. Several other peripherals are connected
to the microcontroller or to the embedded computer such as communication interfaces, and output
devices. The microcontroller is a special purpose computer and its main responsibility is to directly
communicate with the motors and adjust their power levels. Usually, the microcontroller provides
a programming interface to the embedded computer through which a user commands the main
robot functionalities. Other minor functionalities are already embedded in the firmware of
the microcontroller as for example the monitoring of the motors temperature. The embedded
computer is a general purpose machine with much higher computational capabilities. Peripheral
devices like network adapters, monitors, keyboards, cameras etc. are connected to this machine.
The whole system is usually powered by batteries which are located in the robot chassis. Some
robots also use solar surfaces to partially recharge the batteries and delay their exhaustion
\cite{1642356}.
Depending on the batteries storage capacity and the consumption of the components, a robot may
operate from a few minutes to several hours.

\begin{figure}[!ht]
\centering
\includegraphics[height=180pt]{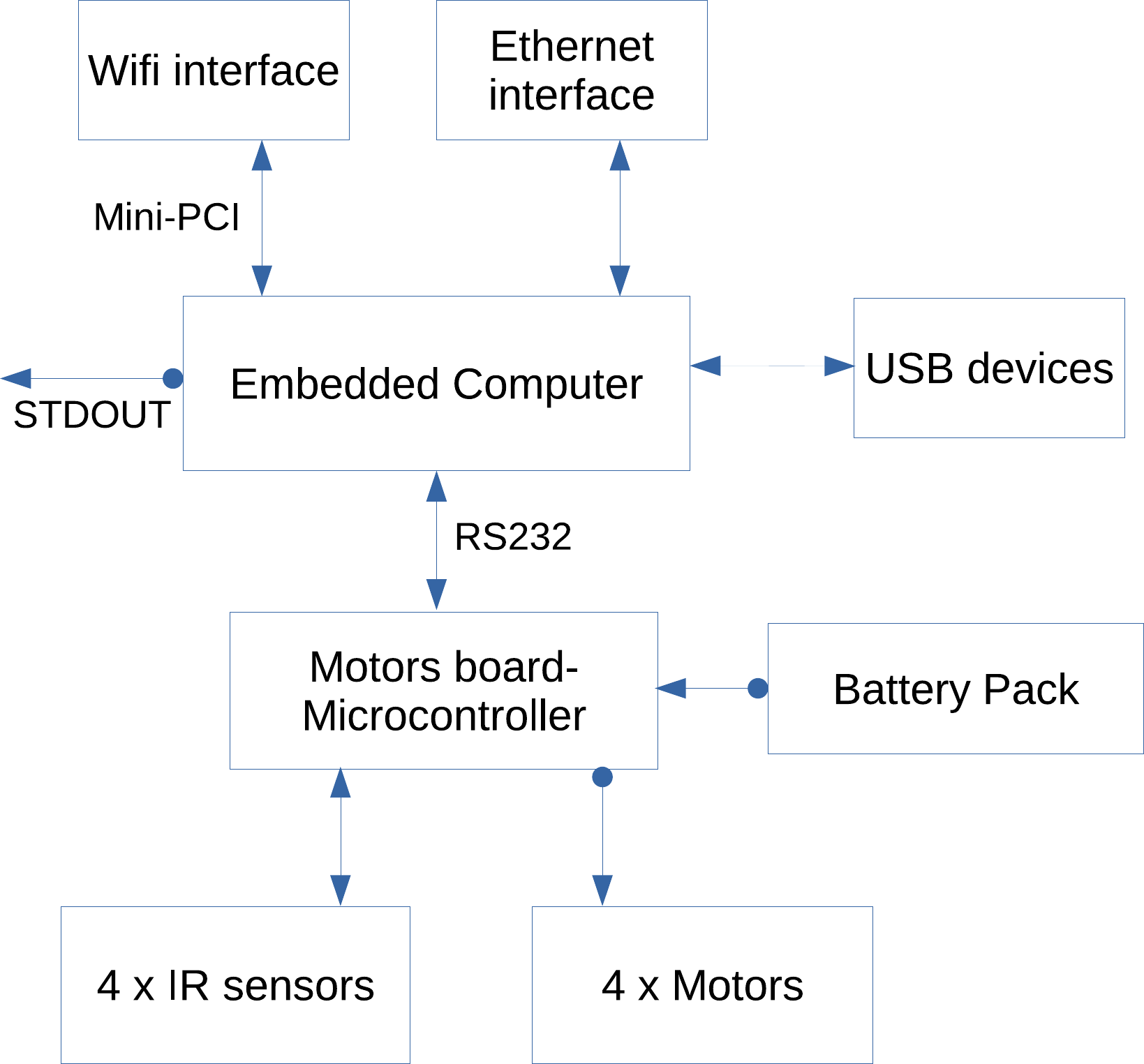}
\caption{Wifibot architecture.}
\label{arch}
\end{figure}

Figure \ref{arch} illustrates an overview of the Wifibot architecture. The embedded computer
plays the role of a motherboard in common computers where all peripherals are connected on it.
A low consumption x86 processing unit and a flash disk are used to reduce power consumption. The
embedded computer communicates with a motor board through a serial port. The motor board plays
the role of the microcontroller and the power regulator. It is responsible of handling low-level
controls of the motors and the IR distance sensors. The power
supply is connected to the motor board through which is distributed to the microcontroller and the
other robot components \cite{wifibotdatasheet}.

\section{Experimental Setup \& Results}

\subsection{Setup}

\begin{figure}[!ht]
\centering
\includegraphics[height=150pt]{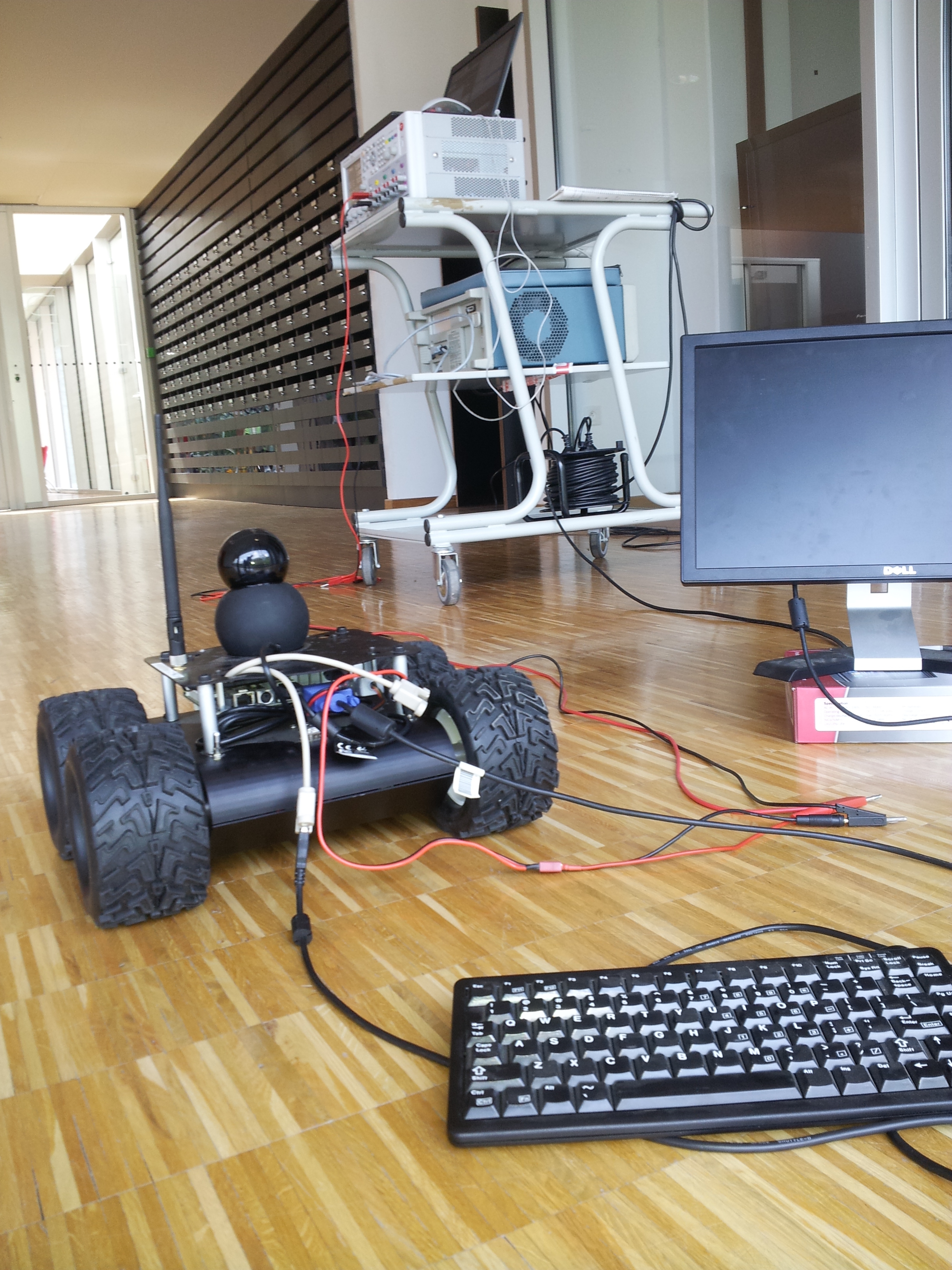}
\caption{The experimental setup.}
\label{exp_setup}
\end{figure}

All the experiments were carried out using a Wifibot mobile robot, a power analyzer connected
to the robot, a keyboard, and a monitor. The third version of a stock Wifibot was used which
weighs approximately 4kg including lithium iron phosphate batteries. The robot's maximum speed
is 0.9 m/s. For the needs of our experiments we bypassed the battery assembly by directly
powering up the robot with the power analyzer. A cable was used to connect the robot with
the power analyzer as it is illustrated in Figure \ref{exp_setup}. The cable was long enough
in order to move the robot many meters away without disconnecting the analyzer. The analyzer
was placed on a cart pushed by a person in case we wanted to travel very long distances. The
commands were given by a USB keyboard and the standard output was displayed on an external
VGA monitor. Both the keyboard and the monitor were detached before the robot starts moving.
A number of scripts was developed to control the robot's maximum traveling distance and speed.
After each experiment we used to reposition the robot to its initial position, reconnecting the keyboard
and the monitor. We exploited the output log files of the power analyzer to calculate the power at
each instance of the experiment. Particularly, one value was taken every 0.1024msec for some
seconds. All the experiments took place on a flat surface of a clean non-slippery parquet-style
floor providing a good grip without spinning while accelerating or moving.

\subsection{Assessing motion power consumption}

In this section, we measure the overall power consumption based on the utilization of the motors.
We compare the results to different robot running modes. We separate the power analysis in
two parts. In the first part, we analyze the power increase caused by acceleration,
while in the second part we model the normal (constant) consumption at a given speed.

The power consumed by a mobile robot is the sum of the power consumed by the motors (mechanical power)
and the power consumed by the embedded devices (see Formula \ref{ptotal}).
The mechanical power (i.e, $P_m$) is used to accelerate and maintain a constant speed. It is expected
to be a linear function which depends on the robot mass $m$, the speed $v$, a ground friction
constant $\mu$, and the acceleration of gravity $g$ \cite{1507454,6265887}.
\begin{equation}
\label{ptotal}
 P_{total} = P_e + P_l + P_m = P_e + P_l + mav + g\mu v
\end{equation}
$P_l$ reports to the power loss due to transformation from electrical to mechanical energy and
$P_e$ is the power consumption of the embedded devices. Assuming that $P_l$ is negligible
compared to the other power consumers, the total power expense is a linear function of
the speed. The robot's mass can, also, affect the energy cost during the acceleration.

\begin{figure}[!ht]
\centering
\subfigure[0.3 m/s]{\includegraphics[height=100pt]{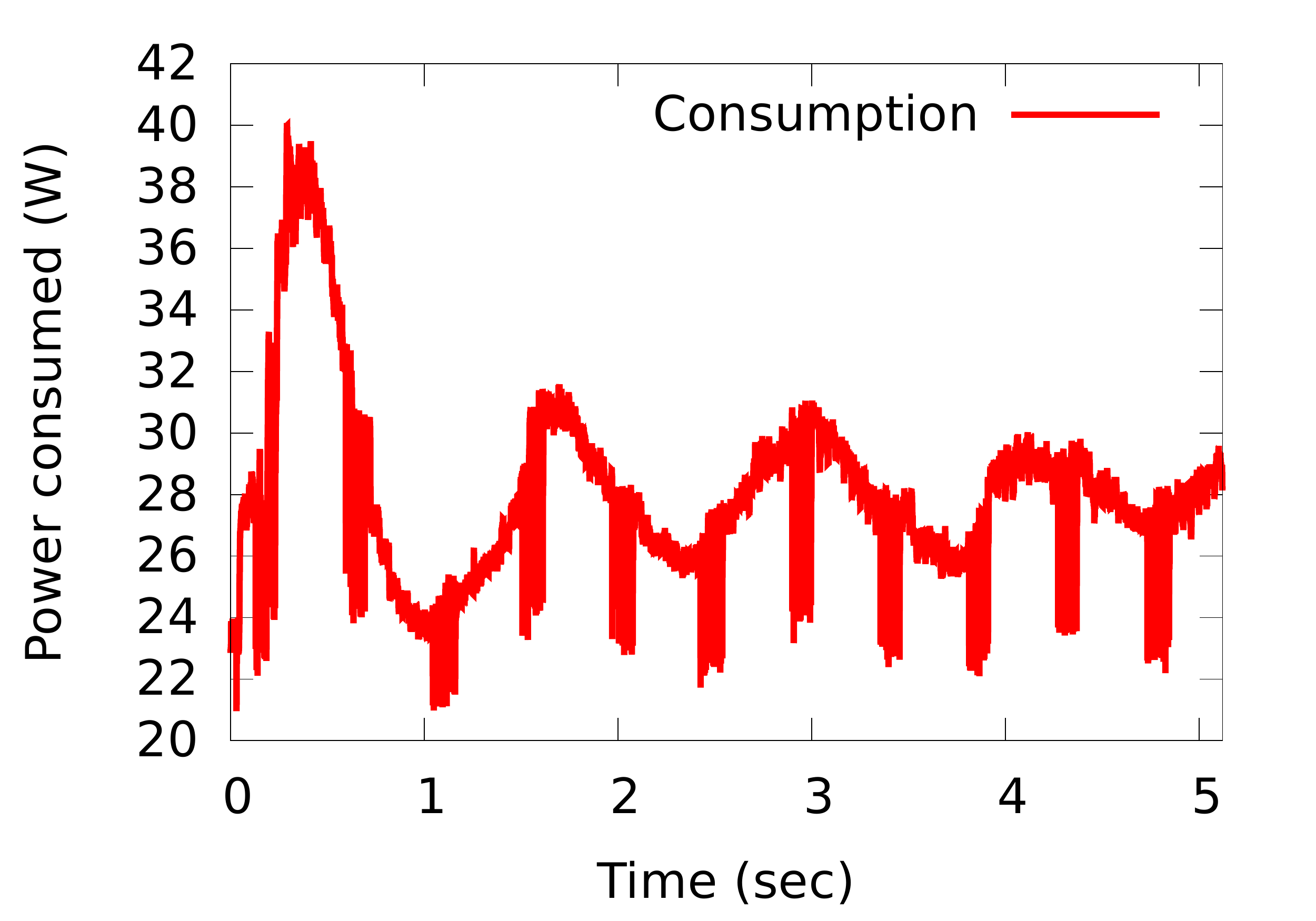}}
\subfigure[0.5 m/s]{\includegraphics[height=100pt]{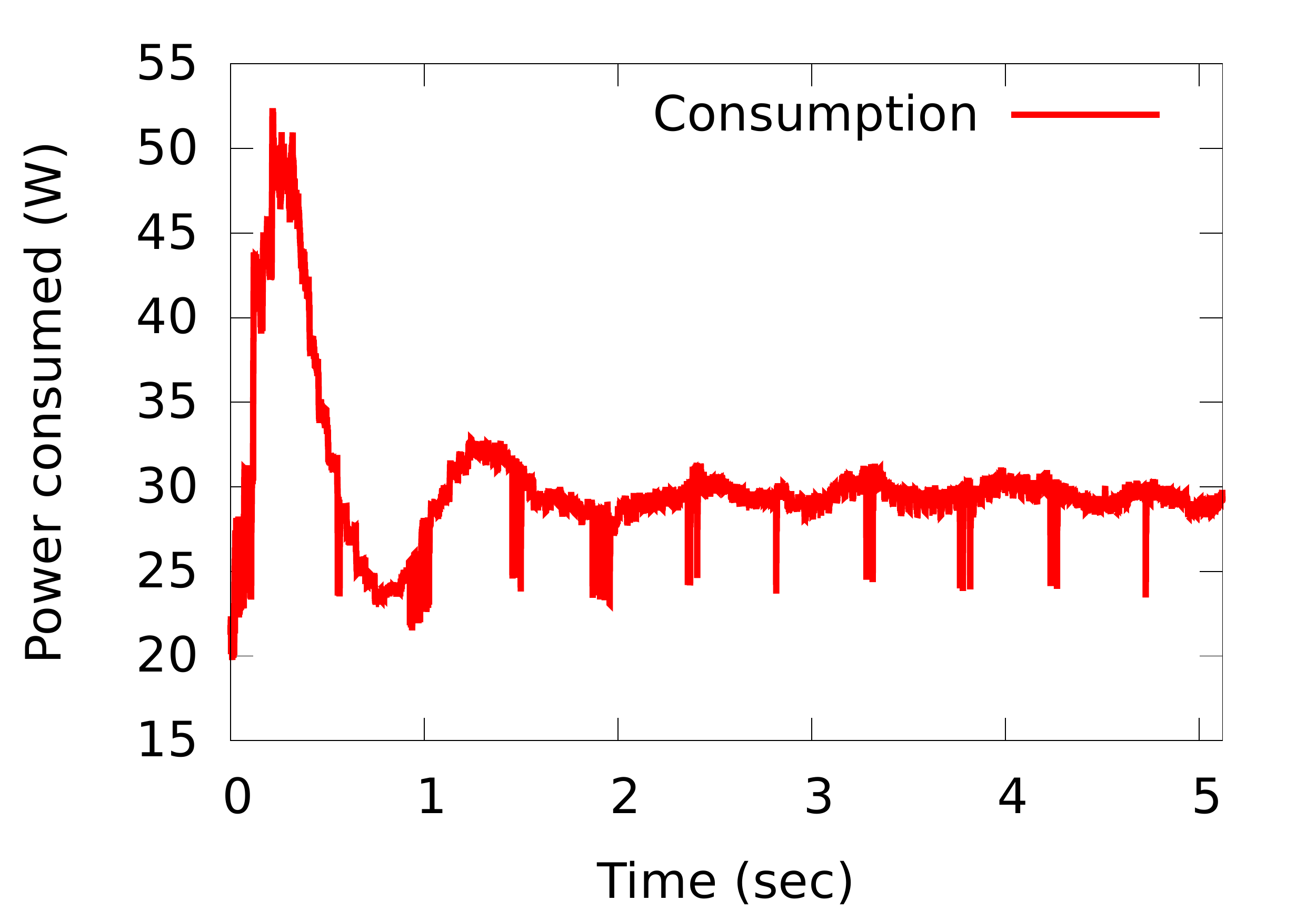}}
\subfigure[0.8 m/s]{\includegraphics[height=100pt]{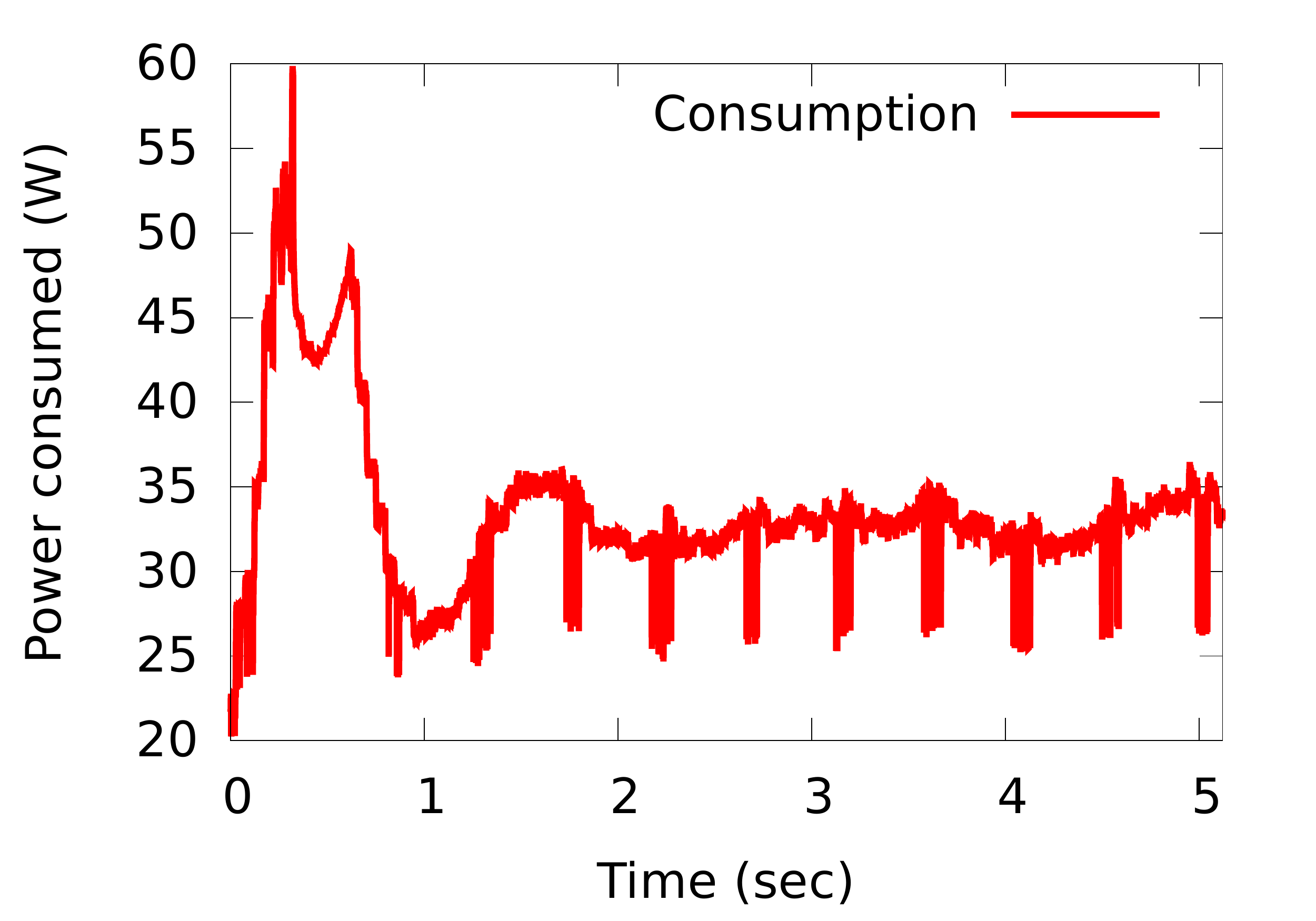}}
\caption{Power consumption for different speeds.}
\label{time_vs_nrg_nr}
\end{figure}

Figure \ref{time_vs_nrg_nr} illustrates the power consumption of the robot for 5 seconds
and different speed levels.
Initially it starts from a stationary position, it accelerates reaching a maximum given speed,
and it keeps moving on a straight line for some seconds.
The displayed consumption also includes the power consumption of the
integrated system which is about 23 Watts (idle mode). All the other unnecessary
components (camera, network adapters etc.) are switched off during all the experiments.
Each experiment was carried out several times (runs) and the average results are presented.
The results show that the power while accelerating may be more than two times higher
than normal consumption.

As we mentioned before, we split the recorded power instances in two parts. The first part concerns
the acceleration power while the second one concerns the power needed to maintain a
constant speed. 
By analyzing the monitoring data and applying the least square method
we conclude that the power used for acceleration is given by:
\begin{equation}
 \label{acc}
 P_{acc_v} = 23.9 + 14.8 v,
\end{equation}
where $v$ is the given maximum speed. This is the average power during the acceleration
phase. 

When the given speed is reached, the power consumption presents a sinusoidal periodicity
which is more clear at lower speeds. 
Each run of an experiment consists of some thousands of values stored in the power analyzer.
Excluding acceleration and computing the average value for each instance of speed, we get
that the power consumption for a given speed $v$ is given by:
\begin{equation}
 \label{no_acc}
 P_{v} = 23.9 + 11.2 v.
\end{equation}

\begin{figure}[!ht]
\centering
\includegraphics[height=120pt]{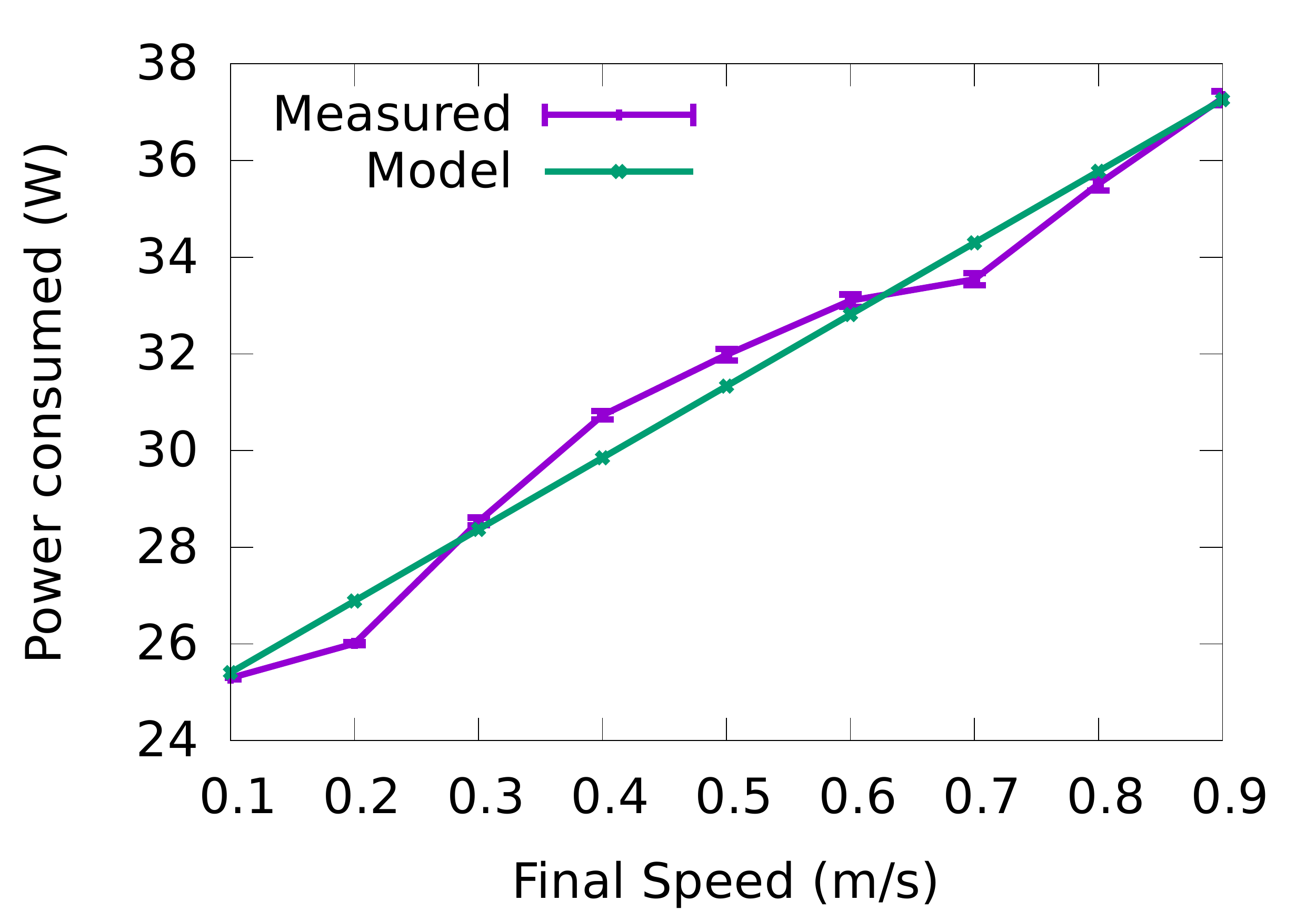}
\includegraphics[height=120pt]{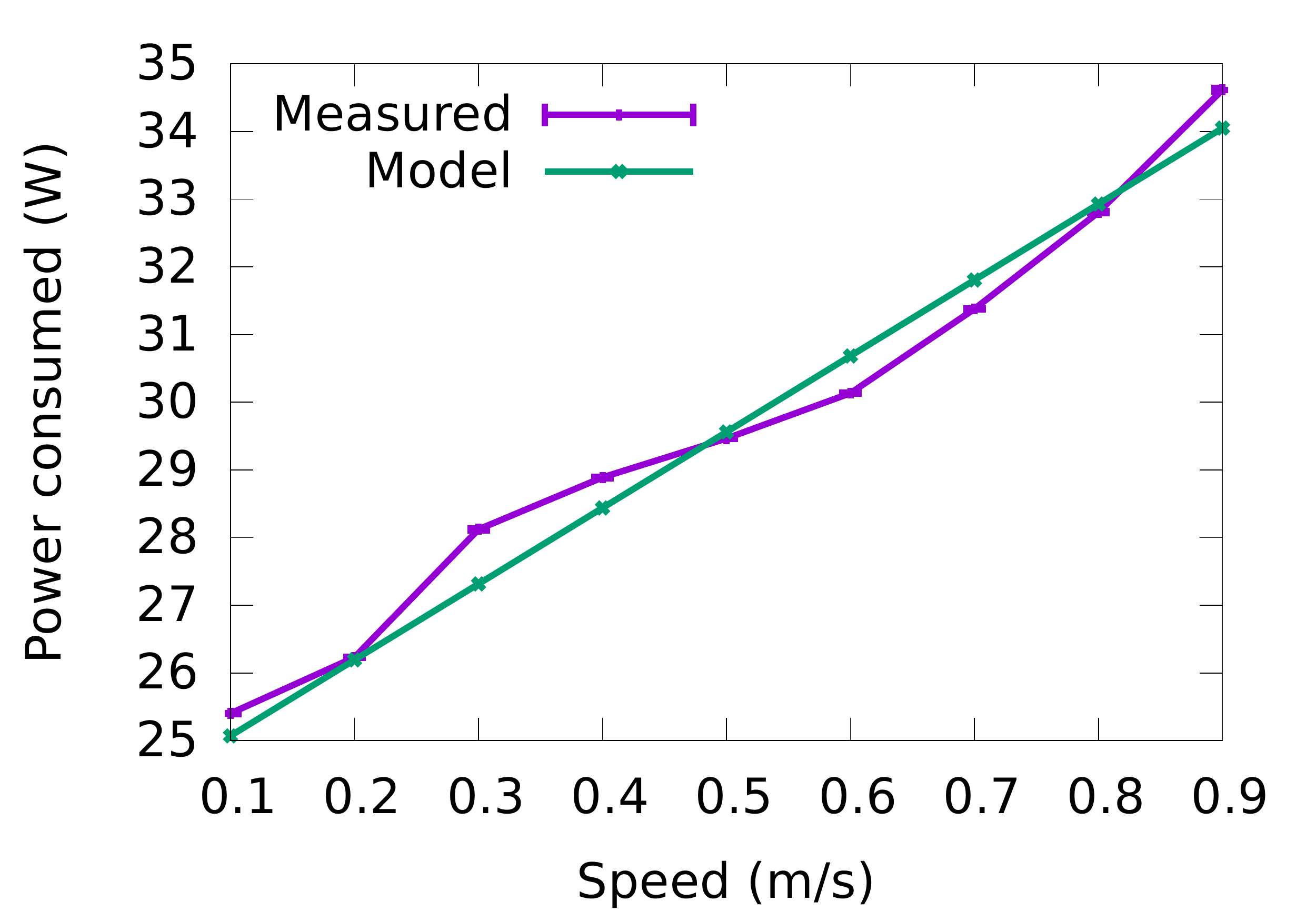}
\caption{Power consumption during acceleration (left) and after acceleration (right)}
\label{speed_vs_nrg_nr}
\end{figure}

Figure \ref{speed_vs_nrg_nr} depicts the power consumption during and after
acceleration. The average experimental values as well as the fitting line are presented.
The average power during acceleration may be up to 9.5\% higher than traveling at constant
speed. This increase in power consumption is high at high speeds and low at low speeds.
Combining Formulas (\ref{acc}) and (\ref{no_acc}), the total energy cost of a single run
is given by:
\begin{equation}
 \label{comb}
 E_v = P_{acc_v}t_{acc_v} + P_v(t - t_{acc_v}) = 3.6vt_{acc_v} + P_vt,
\end{equation}
where $t$ and $t_{acc_v}$ is the total moving time and the acceleration time respectively.
Since after acceleration, the robot performs a linear motion, Formula (\ref{comb}) can
be also written as:
\begin{equation}
 \label{comb_s}
 E_v = P_{acc_v}t_{acc_v} + P_v\frac{s-s_{acc_v}}{v},
\end{equation}
where $s$ and $s_{acc_v}$ is the total traveling distance and the distance traveled during
acceleration respectively. From Formulas (\ref{comb}) and (\ref{comb_s}) we can conclude 
that the extra energy cost due to acceleration is not negligible unless low-speed or very
long runs are performed.

\begin{figure}[!ht]
\centering
\includegraphics[height=120pt]{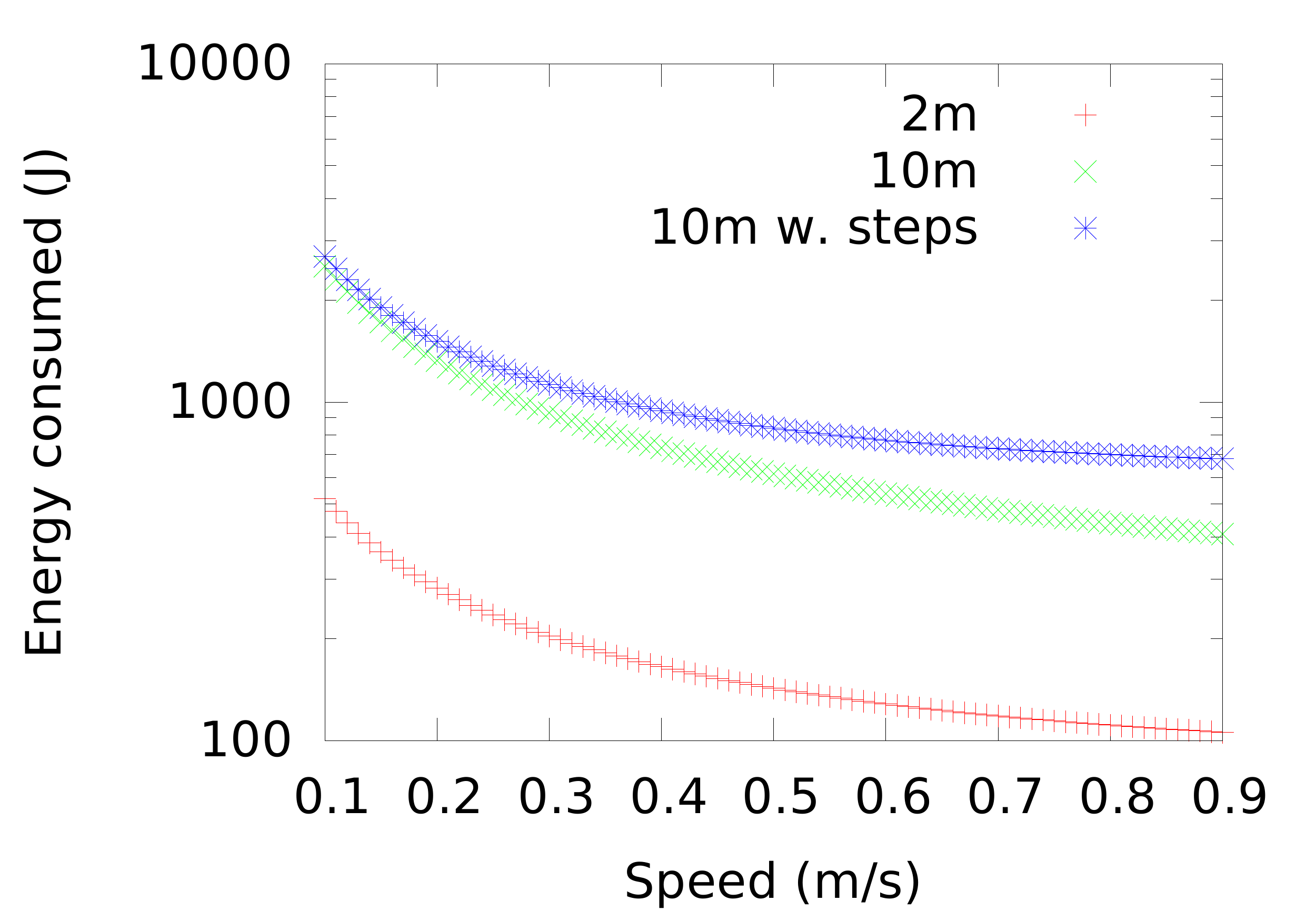}\\
\caption{Motion consumption for different speeds while traveling 2 or 10 meters}
\label{nrg-spd-dst}
\end{figure}

To clarify this statement we conduct another experiment where we vary the speed and the
traveling step (distance traveled before the robot stops). The results are presented in Figure
\ref{nrg-spd-dst}. We can observe that the energy is higher for lower speeds since the time
to travel the distance is longer. When the distance is 10 meters, we distinguish an extra
case to assess the effect of acceleration in total consumption. In this case, the robot
travels the distance with equal steps of 1 meter each. The robot stops at the end of each
step. The results show that the energy consumption in high speeds is up to 66\% higher
than completing the task in one step.

\subsection{Activating other components}

Table \ref{table1} summarizes the global power while utilizing a number of robot components and
compares the results to the consumption when the robots moves at maximum speed. In particular,
in addition to the motion power consumption, we measure the
power consumption when (a) the robot is idle, (b) it is transmitting data using the highest
possible bandwidth of the wifi interface (802.11g), (c) a video is taken using the integrated webcam
(jpeg format), and (d) a CPU core is stressed with maximum load. The power values
in parenthesis are the net values without including the idle power consumption.

The results reveal that more than 50\% of the overall power is consumed when the robot
stays idle. This is actually the power consumption of the embedded system and equals
to the minimum possible consumption of the robot while it is in power on mode
\footnote{some energy savings may be achieved if some of the computer components, like cpu cores, usb,
infrared etc, are turned off}. The embedded system power consumption is apparently high,
but is similar to that of other mobile robots, like the one tested in \cite{1507454}.
The motion consumption comes second with 25\% of the
overall power while the network consumption, the webcam usage and the CPU stress test
are in third, fourth and fifth place respectively.

\begin{table}
\begin{center}
\begin{small}
\begin{tabular}{|c|c|c|c|}
\hline
\textbf{Function} & \textbf{Power (W)} & \textbf{Max. Motion (\%)} & \textbf{Overall (\%)}\\
\hline
Max. Motion & 33.98 (10.68) & 0 & 24.85\\
Idle & 23.3 & 118.16 & 54.21\\
Network & 27.7 (4.4) & -58.8 & 10.24\\
Video & 25.8 (2.5) & -76.59 & 5.82\\
CPU & 25.4 (2.1) & -80.34 & 4.89\\
\hline
\end{tabular}
\end{small}
\end{center}
\caption{The power consumption while activating different robot devices.}
\label{table1}
\end{table}

\section{The communication effect on the operation time}

In our measurements wifi transmission consumes about 10\% of the total power consumption.
Other experimental measurements of wifi interfaces have shown that power consumption while
transmitting or receiving costs only some nanowatts per bit \cite{feeney,80211n}. However,
usually network devices adapt their transmission power to the trasmission range in order
to conserve some energy. We take advantage of this capability and we study
whether a reduced distance between the transmitter and the receiver could lead, in long-term,
to lower total power consumption and, thus, to longer operation times.

\begin{figure}[!ht]
\centering
\includegraphics[height=80pt]{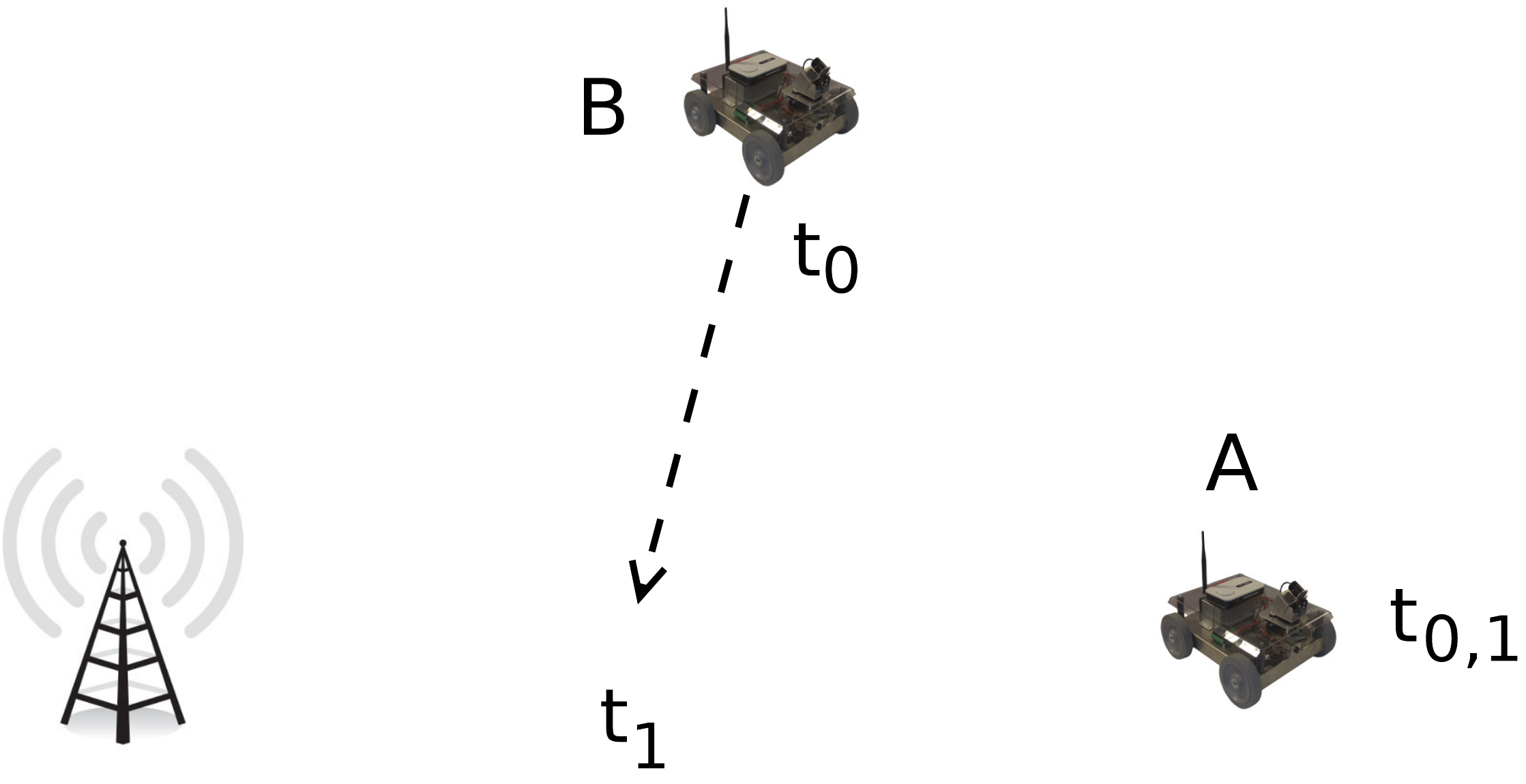}\\
\caption{Robot network scenario.}
\label{scenario}
\end{figure}

We consider a simple scenario where a stationary robot, which is used
for monitoring and communicates with a base station, decreases its transmission power
forwarding its monitoring data through a nearby relay robot. Figure \ref{scenario}
illustrates the examined scenario. We assume that the stationary robot, annotated with
``A'', is located far from the base station consuming a lot of energy for transmissions.
Robot B is located close to A. The two robots are in the communication range of each
other. We move robot B towards a midpoint between A and the base station, operating as
relay node, an action which we would wish to decrease the consumption of monitoring
robot and, thus, would increase the monitoring time.

We assume that B's movement starts at $t_0$ and stops at $t_1$. During B's movement,
``A'' communicates directly with the base station and for this period of time the
total amount of energy it consumes is equal to the following sum:
\begin{equation}
\label{A0}
E_{A_0} = P_{tx}^{A_0}\ \delta\ t^{0 \rightarrow 1} + P_i\ t^{0 \rightarrow 1} + P_s\ \delta\ t^{0 \rightarrow 1},
\end{equation}
where $P_{tx}^{A_0}$ is the power consumption due to data transmission, $P_s$ is
the sensing power consumption, and $P_i$ is the power consumption of the
embedded system. $\delta$ is the amount of data transmitted per time unit.
When ``B'' reaches its destination, it starts operating as relay node. From
this moment its energy consumption is computed as follows:
\begin{equation}
E_{B_1} = P_{tx}^{B_1}\ \delta\ t^{1 \rightarrow \max} + P_i\ t^{1 \rightarrow \max} + P_{rx}\ \delta\ t^{1 \rightarrow \max},
\end{equation}
where $P_{rx}$ is the power consumption due to data reception.

$P_{tx}$ is a non-fixed value which depends on transmitter's power. The higher
the power, the longer the transmission distance. $P_{rx}$, $P_i$, and $P_s$ are
fixed.

Assuming an even energy reduction of the initial battery storage, the lifetime of
the network is computed by dividing the initial energy of the robot at $t_0$ by
its energy consumption per time unit. The energy consumption per time unit is
actually the average power consumption and it is annotated with $P$.

The network lifetime depends on the initial energy of the two robots at $t_1$
and their power consumptions. According to the value of these parameters any
of the two robots may die first interrupting the network activity. If robot
A dies first then the network lifetime $L$ is given by:
\begin{equation}
\label{L}
L = t_B^{0 \rightarrow 1} + \displaystyle \frac{E_{0_A} - E_A^{0 \rightarrow 1}}{P_{A_1}},
\end{equation}
where $t_B^{0 \rightarrow 1}$ is the moving time of B and $E_A^{0 \rightarrow 1}$
is the energy consumed by A while B is moving.
$E_A^{0 \rightarrow 1}$ is computed by choosing a transmission power $P_{tx}$ capable of
transmitting data to the base station with a very low bit error rate \cite{1201163}. 

If robot B dies first, the network lifetime $L'$ is computed by the following sum:
\begin{equation}
\label{L_}
L' = t_B^{0 \rightarrow 1} + \displaystyle \frac{E_{0_B} - E_m^{0 \rightarrow 1}}{P_{B_1}} +
\displaystyle \frac{E_{0_A} - E_A^{0 \rightarrow 1} - E_A^{1 \rightarrow 2}}{P_{A_0}}.
\end{equation}
The first term is the moving time of B, the second term is the lifetime of B, and the
third term is the time ``A'' can directly communicate with the base station after B's death.
If we assume that ``B'' dies at moment $t_2$, then $E_A^{1 \rightarrow 2}$ is the energy
reduction of robot A during the period between $t_1$ and $t_2$. $E_A^{1 \rightarrow 2}$
is computed by Formula (\ref{EA2}).

\begin{equation}
\label{EA2}
E_A^{1 \rightarrow 2} = \displaystyle \frac{E_{0_B} - E_m^{0 \rightarrow 1}}{P_{B_1}} P_{A_1}.
\end{equation}

We compute the lifetime improvement over the initial scenario where B is not used as relay
node at all. In this case the network lifetime would be equal to $\frac{E_{0_A}}{P_{A_0}}$.
We consider an initial battery capacity as it is specified in Wifibot V3 specifications
\cite{wifibots}, constant data bitrates from 10 to 100Mbit/sec, and a IEEE 802.11n power
consumption model for $P_{tx}$ and $P_{rx}$ as it is experimentally measured in \cite{80211n}.
$P_{tx}$ and $P_{rx}$ are measured in nW/bit and they depend on the actual data bitrate
(i.e., $\delta$). The higher the data rate, the lower the value of $P_{tx}$ and $P_{rx}$.
We summarize the parameters used in Table \ref{parameters}.

\begin{table}
\label{parameters}
\begin{center}
\begin{tabular}{|c|c|}
\hline
\textbf{Parameter} & \textbf{Values}\\
\hline
$E_{0_{A,B}}$ (when not variable) & 480 kJ\\
$E_{0_{A,B}}$ (when variable) & 50 to 450 kJ\\
$\delta$ (when not variable) & 100Mbit/sec\\
$\delta$ (when variable) & 10, 54, 100Mbit/sec\\
$P_{tx}$ (high trans. power)& 95, 24, 12nW/bit\\
$P_{tx}$ (low trans. power)& 9.5, 2.4, 1.2nW/bit\\
$P_{rx}$ & 70, 18, 10nW/bit\\
$P_s$ & 100nW/bit\\
travel. dist. (when variable) & 2 to 10m\\
travel. dist. (when not variable) & 10m\\
$E_m$ & by Formula (\ref{comb_s})\\
speed & 0.9 m/sec\\
\hline
\end{tabular}
\caption{Parameters and values used.}
\end{center}
\end{table}

\begin{figure}[!ht]
\centering
\subfigure[Lifetime improvement per initial energy.]{\includegraphics[height=120pt]{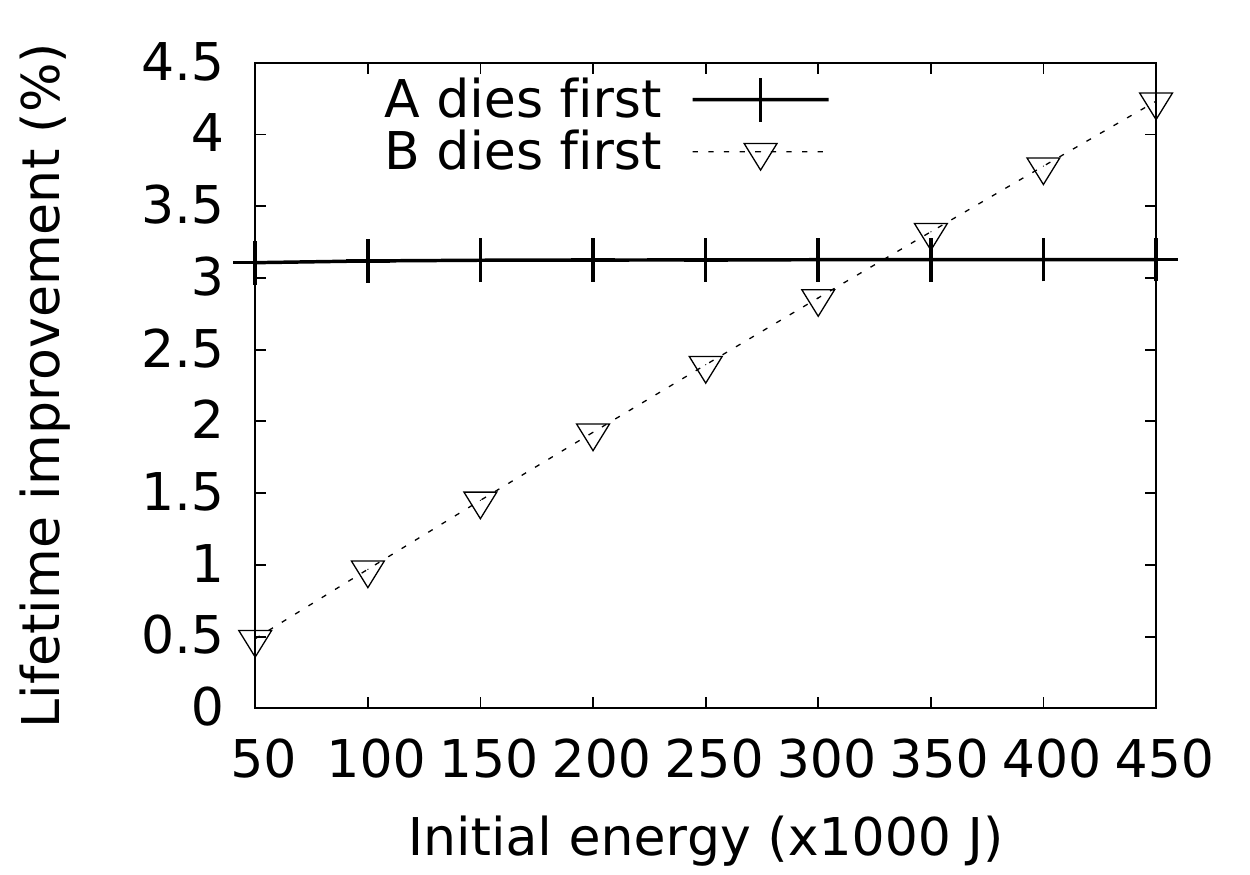}}
\subfigure[Lifetime improvement per traveling distance.]{\includegraphics[height=120pt]{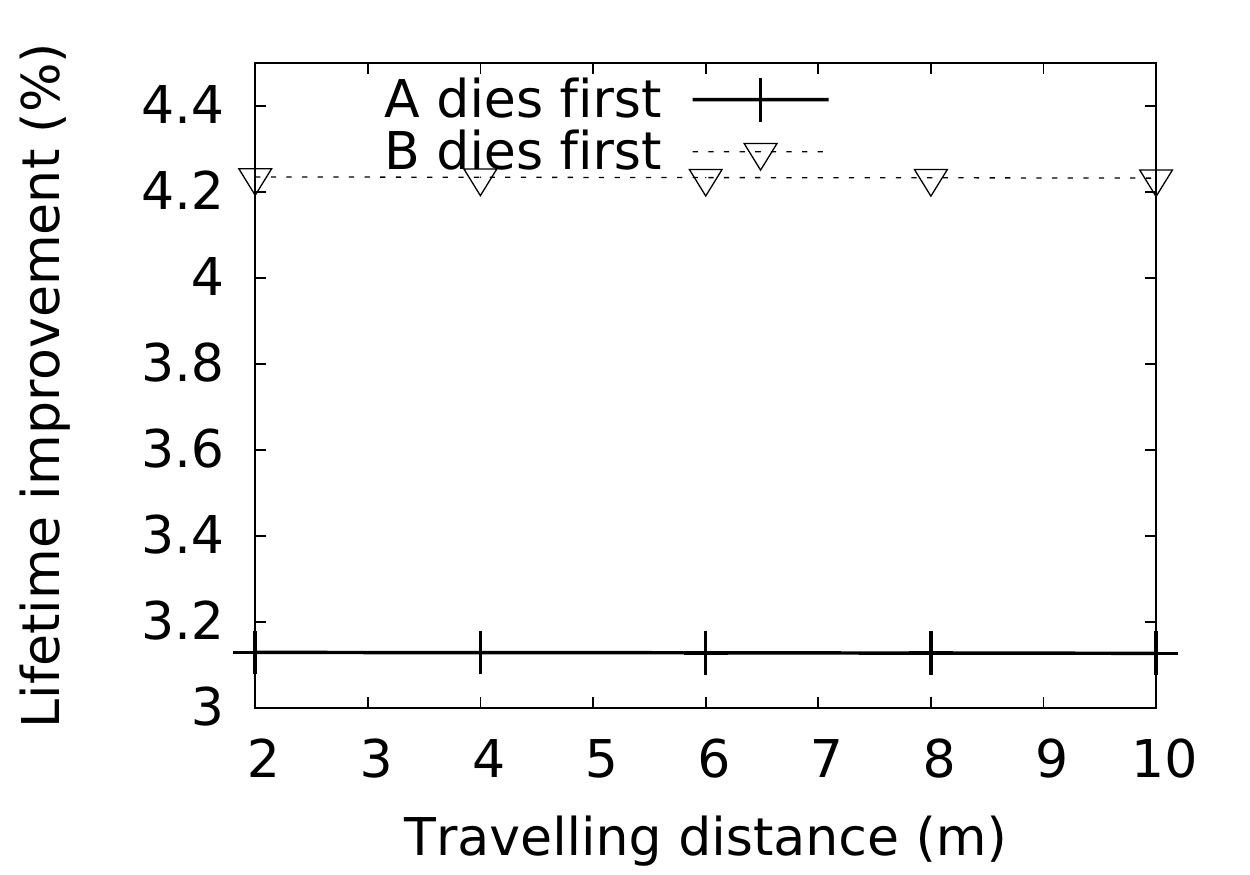}}
\subfigure[Lifetime improvement per data bitrate.]{\includegraphics[height=120pt]{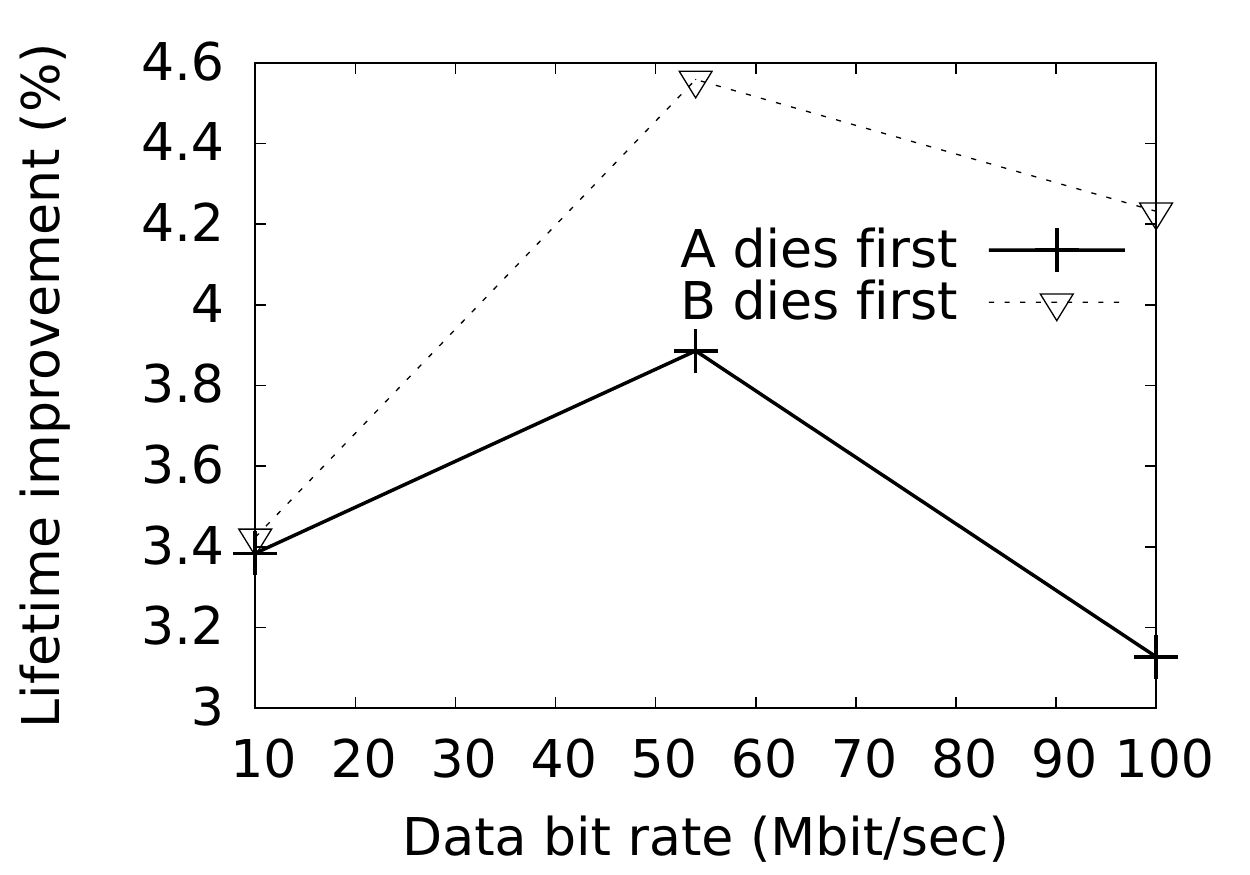}}
\caption{Lifetime improvement when another robot is used as data relay.}
\label{lft}
\end{figure}

We vary the initial energy capacity, the traveling distance and the data bitrate of either
robot A or B for Formulas (\ref{L}) and (\ref{L_}) respectively. According to these formulas,
A or B may die first. For this reason, we distinguish two cases; in the first one, we keep
constant the initial energy capacity of B, varying the initial capacity of A, while in the
second one we do the opposite. The X axis of Figure \ref{lft}a represents the initial energy
of either A or B, when it is not constant.

The results are presented
in Figure \ref{lft} and show minor lifetime improvements ranging from 0.5 to 4.5\%. 
One could say that the operation time cannot improve a lot since the specific embedded system
consumes a lot of power. However, our simulations revealed a higher, but still low improvement (up to 7.6\%),
considering an embedded system with the half of the power consumption of our specific
robot. We omit these results since their trend is the same as in Figure \ref{lft};
they only differ in terms of absolute values.
The results,
also, show that the lifetime presents a linear increase as the initial energy of robot B is growing.
It means that considering higher energy capacities, the lifetime could be further
improved.

\section{Conclusion}

In this paper, we modeled the power consumption of several running modes of a mobile robot called Wifibot.
We conducted a number of experiments giving attention to specific motor running modes,
such as acceleration and constant speeds. Based on the consumption of these running modes, we built
models for acceleration and constant speeds. The results showed that if the robot stops frequently,
the energy cost increases up to 66\% mainly due to accelerations. Moreover, high energy efficiency
is achieved at higher speeds.
Even if these results are based on the specific robot, we believe that acceleration is, generally,
an operation with high energy cost and, thus, its frequent use shortens the operation time.
How much it affects performance depends on motors characteristics as well as on robot payload.

We extended our approach to a simulated experiment involving two robots that perform communication
with a distant station. We examine whether a shorter communication link, and thus lower power
consumption, could lead to longer operation times. We considered a simple scenario where a robot
is moved between a distant node which is used for monitoring and a base station. The two robots
adjust their communication ranges to shorter radius to conserve energy. The simulation results show
that low operation time savings can be achieved for the specific robot. However, equations show
that this technique could lead to better results if other robots with lower consumption of the embedded
system or/and high energy storage units were used.

\section*{Acknowledgements}
The authors gratefully acknowledge the comments received from Roman Igual
during the experimentation setup and the help he provided using the IRCICA
telecommunication platform (http://www.ircica.univ-lille1.fr).

\bibliographystyle{plain}
\bibliography{robot_nrg}

\end{document}